\documentclass[conference]{IEEEtran}
\usepackage{cite}

\ifCLASSINFOpdf
  \usepackage[pdftex]{graphicx}
\else
  \usepackage[dvips]{graphicx}
\fi
\usepackage{amsmath,amsthm,amscd,amssymb,latexsym}
\usepackage{url}

\usepackage{bm}
\usepackage[ruled,vlined,linesnumbered]{algorithm2e}
\usepackage{comment}

\begin{document}
%
\title{Transfer Learning for Endoscopic Image Classification}

\author{
Shoji Sonoyama, Toru Tamaki,\\
Tsubasa Hirakawa, Bisser Raytchev,\\
Kazufumi Kaneda, Tetsushi Koide\\
Hiroshima University\\
{\tt\small tamaki@hiroshima-u.ac.jp}
\and
Shigeto Yoshida, Hiroshi Mieno\\
Hiroshima General Hospital of West Japan Railway Company\\
Shinji Tanaka\\
Hiroshima University Hospital
}


%


\maketitle

\begin{abstract}
In this paper we propose a method for transfer learning of 
endoscopic images. For transferring between features obtained from images taken by different (old and new) endoscopes, we extend the
Max--Margin Domain Transfer (MMDT) proposed by Hoffman et al.
in order to use $L_2$ distance constraints as regularization, called Max--Margin Domain Transfer with $L_2$ Distance Constraints (MMDTL2).
Furthermore, we develop the dual formulation of the optimization problem in order to reduce the computation cost.
Experimental results demonstrate that the proposed MMDTL2 outperforms MMDT
for real data sets taken by different endoscopes.
\end{abstract}


%
\IEEEpeerreviewmaketitle

\section{Introduction}

Nowadays in many hospitals, an endoscopic examination (colonoscopy) using Narrow Band Imaging (NBI) system is
widely performed to diagnose colorectal cancer \cite{Tanaka2006},
which is one of the major cause of cancer death \cite{CancerResearchUK}.
During examinations, endoscopists observe and examine a polyp 
based on its visual appearance such as an NBI magnification findings \cite{Kanao2009,Oba2010}.
To support diagnosis during examinations,
a computer-aided diagnosis system based on the texture appearances of polyps would be helpful,
and therefore patch-based classification methods for endoscopic images have been proposed
\cite{Hafner2010a,Hafner2010b,Kwitt2010,Gross2009,Stehle2009,Tischendorf2010,Tamaki2013MedIA}.

The problem we address in this paper is the inconsistency between training and testing images \cite{sonoyama2015transfer}.
As other usual machine learning approaches, training of classifiers assumes that
the distributions of features extracted from both training and testing image datasets are the same.
However, different endoscopies may be used to collect training and testing datasets, leading the assumption to be violated.
One reason is due to the rapid development of medical devices (endoscopies in our case),
and hospitals could introduce new endoscopes at a certain point in time after training images were taken.
Another reason is that a dataset is constructed for training classifiers with a training data set collected by a certain type of endoscope in one hospital,
while another hospital could want to use the classifiers for images taken by a different endoscope.
In general, such kind of inconsistency could lead to a deterioration of the classification performance,
hence collecting new images for a new training dataset may be necessary in general.
It is however not the case with our medical images: it is impractical to collect a enough set of images for all types and manufactures of endoscopes.

\begin{figure}[t]
 \begin{minipage}{0.45\linewidth}
  \centering
   \includegraphics[width=\linewidth]{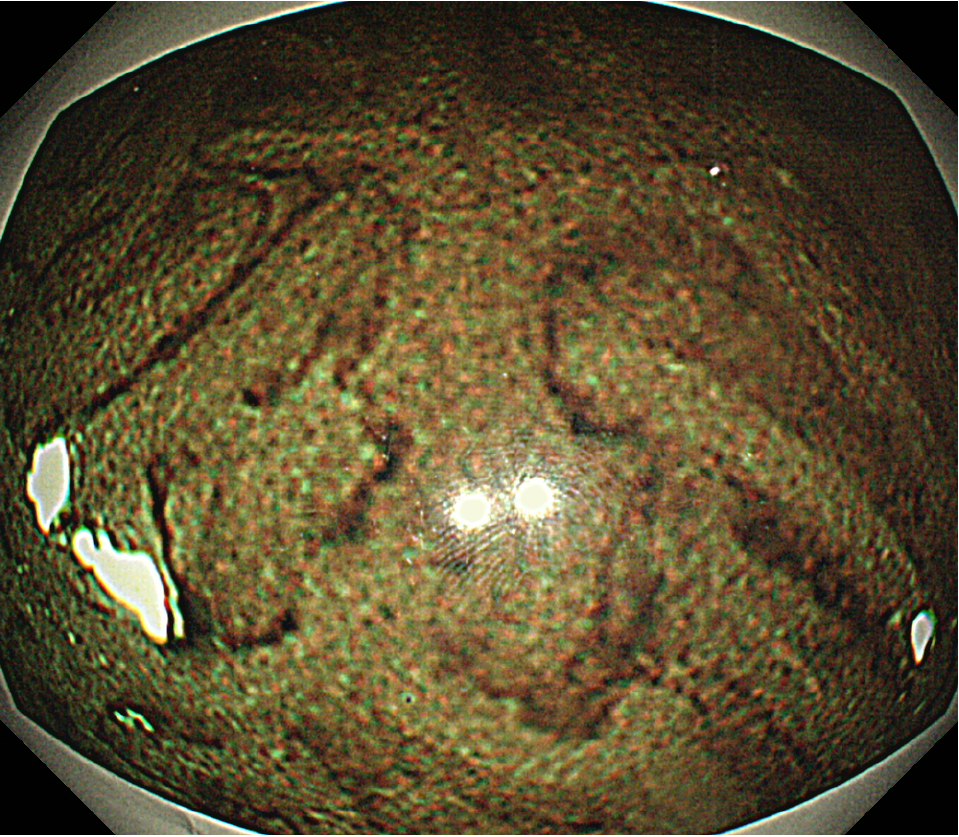}
  (a)\\
 \end{minipage}
 \hfill
 \begin{minipage}{0.45\linewidth}
  \centering
   \includegraphics[width=\linewidth]{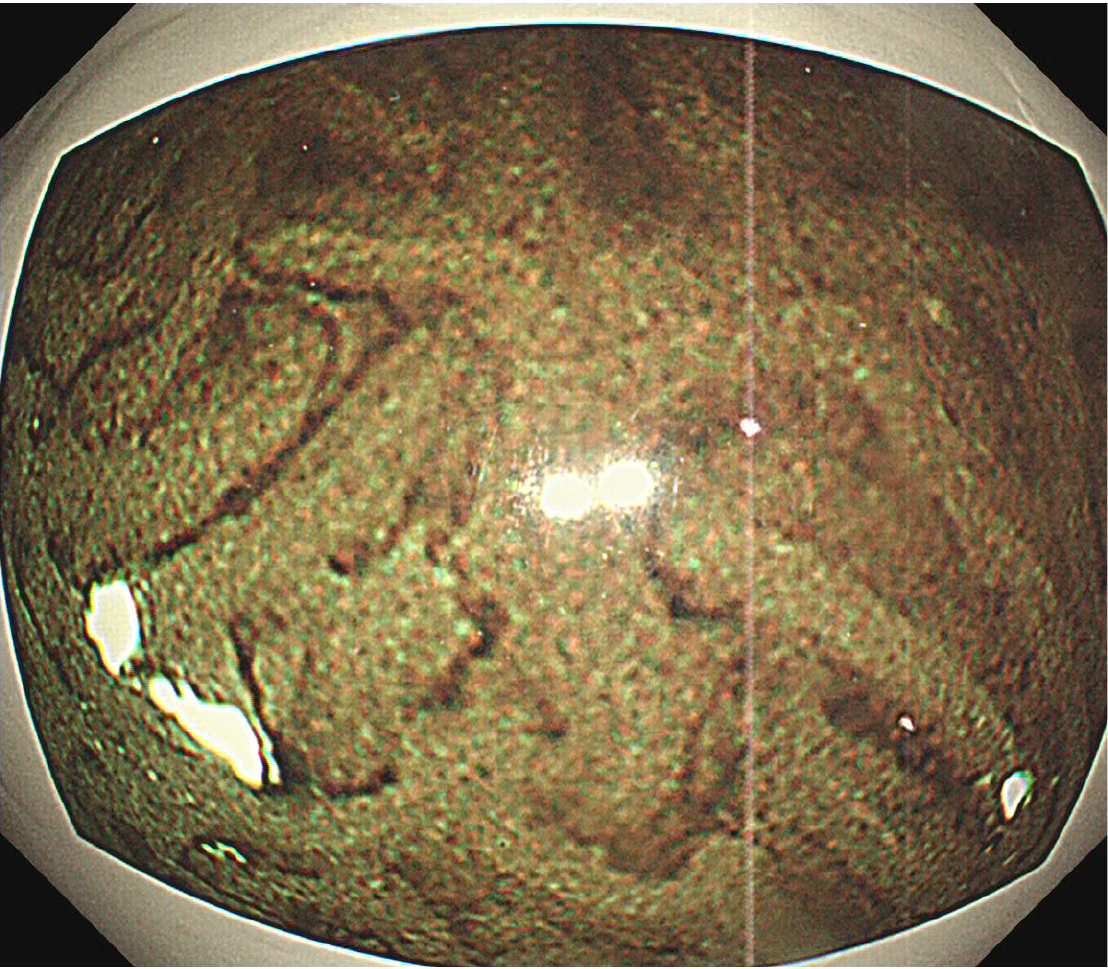}
  (b)\\
 \end{minipage}
 \caption{An example of appearance difference of different endoscope systems.
(a) An image taken by an older system
(video system center: Olympus EVIS LUCERA CV-260,
endoscope: Olympus OLYMPUS EVIS LUCERA CF-H260AZL/I \cite{EVIS_LUCERA_H260AZL}).
(b) An image of the same scene taken by an newer system
(video system center: Olympus EVIS LUCERA ELITE CV-290,
endoscope: OLYMPUS EVIS LUCERA ELITE CF-HQ290ZL/I \cite{EVIS_LUCERA_ELITE_HQ290Z}).
}
 \label{fig:appearance}
\end{figure}

Figure \ref{fig:appearance} shows an example of the difference between appearance of texture taken by different endoscope systems.
These images are the same scene of a printed sheet of a colorectal polyp image,
but taken by different endoscope systems, at almost the same distance to the sheet from the endoscopes.
Even for the same manufacture (Olympus) and the same modality (NBI),
images differ to each other in resolutions, image quality and sharpness, brightness, viewing angle, and so on.
This kind of difference may affect the classification performance.

In order to tackle this problem,
we have proposed a method \cite{sonoyama2015transfer} based on transfer learning \cite{pan2010survey,raina2007self,dai2007boosting,silver2005nips}
for estimating a transformation matrix between feature vectors of training and testing data sets taken by different (old and new) devices.
In this prior work, we formulated the problem as a constraint optimization, and developed an algorithm to estimate the linear transformation.
The problem however is that corresponding data sets are required, in other words, each of test image (taken by a new device) must have a corresponding training image (taken by an old device), and moreover these images must capture the same polyp in order to estimate the linear transformation.
This restriction is very strong and even impractical.

In this paper, we propose another method for the task, but without image-by-image correspondences between training and test data sets.
More specifically, we extend Max--Margin Domain Transfer (MMDT) \cite{Hoffman_ICLR2013}
in order to use $L_2$ distance constraints as a regularization, called Max--Margin Domain Transfer with $L_2$ Distance Constraints (MMDTL2).
This extension is our first contribution, and the second contribution is the derivation of the dual problem to the original primal problem,
which greatly reduce the computation cost.
Experimental results with real endoscopic images show that the our proposed method, MMDTL2, outperforms the previous method, MMDT.

\begin{figure}[t]
  \centering
  \includegraphics[width=\linewidth]{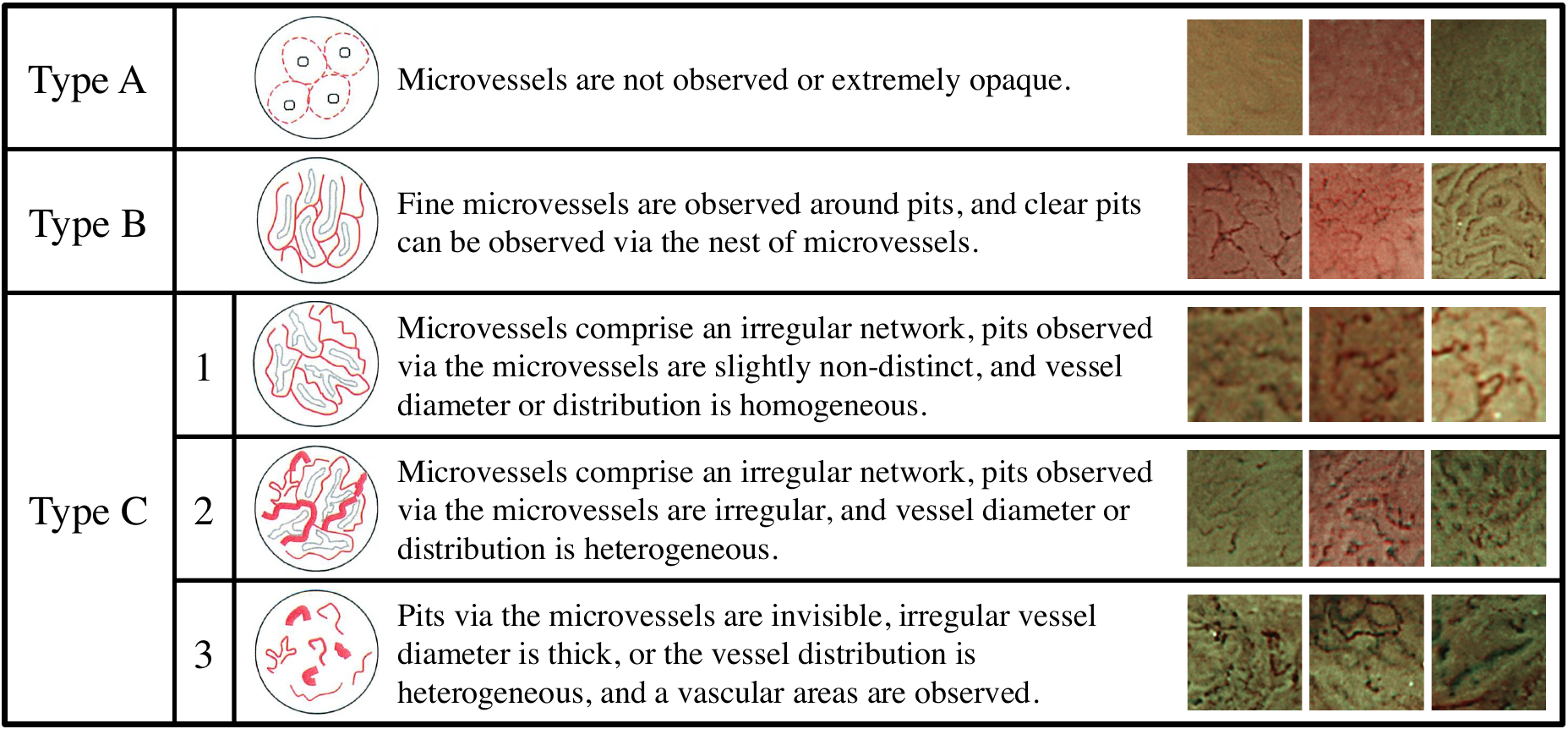}
  \caption{NBI magnification findings \cite{kanao2009narrow}.}
  \label{fig:nbi_magnification}
\end{figure}

Here we formulate the problem setting.
We are given a data set of images $I_n^s \ (n=1,\ldots,n_S)$ taken by an older device, and corresponding feature vectors $\bm{x}_n^s \in R^{M_S}$
and labels $y_n^s \in \{1,\ldots,K\}$.
By following the terms of domain adaptation, 
we call this data set ``source'' (or source domain) because it is the set or domain from which features are transferred.
Then another data set of images
$I_n^t \ (n=1,\ldots,n_T)$ with corresponding features 
$\bm{x}_n^t \in R^{M_T}$ and labels $y_n^t  \in \{1,\ldots,K\}$. We call this ``target'' (or target domain) because this is the destination to which features are transferred.
If the target dataset was large enough so that we could train classifiers by using the target dataset as a training set. In our task we assume that 
the target data set is small compared to the source data set
because of a usual setting that images are collected before changing the image device but still we want to have a good performance, as we are currently facing.

\section{Previous work: MMDT}

A method for transfer learning proposed by Hoffman et al. \cite{Hoffman_ICLR2013}, called MMDT, does the estimation of the linear transformation $\bm{W} \in R^{M_S\times (M_T+1)}$ 
between source features $\bm{x}_i^s$ and target features $\bm{x}_j^t$
and the learning parameters $\bm{\theta}_k \in R^{M_S}$ of $k$-th binary SVM (for a $K$ class problem) at the same time
by solving the following optimization problem (note that the following form is obtained after introducing slack variables $\xi$).
\begin{equation}
\begin{split}
  \min_{ {\bm{W}, \bm{\theta}_k,b_k,} \atop {\xi^s_{i,k}, \xi^t_{j,k} } } \quad 
  & \frac{1}{2}  \|\bm{W}\|_F^2 + {} \\
  & \sum_{k=1}^K  
  \Biggr\{
   \frac{1}{2} \|\bm{\theta}_k\|_2^2  
   + C_S \sum_{i=1}^{n_S} \xi_{i,k}^s 
   + C_T \sum_{j=1}^{n_T} \xi_{j,k}^t 
  \Biggl\} \\
  \mathrm{s.t.} \quad 
  & y_{i,k}^s (\bm{\theta}_k^T \bm{x}_i^s + b_k) \geq 1 - \xi_{i,k}^s \\
  & y_{j,k}^t (\bm{\theta}_k^T \bm{W} \left({\bm{x}_j^t \atop 1}\right) + b_k) \geq 1 - \xi_{j,k}^t \\
  & \xi_{i,k}^s \geq 0, \quad \xi_{j,k}^t \geq 0,
  \end{split}
  \label{eq:MMDT-org}
\end{equation}
where $\bm{\theta}_k$ and $b_k$ are the normal and bias of the hyper plane of a binary SVM classifier for $k$-th class, and $y_{i,k} = \delta(y_i, k)$.
They solve the optimization problem by alternatively estimating $\bm{\theta}$ and $\bm{W}$.
\begin{align}
  \min_{ \bm{\theta}_k,b_k, \xi^s_{i,k}, \xi^t_{j,k}} \
    & \sum_{k=1}^K
      \Biggr\{
       \frac{1}{2} \|\bm{\theta}_k\|_2^2  
        + C_S \sum_{i=1}^{n_S} \xi_{i,k}^s 
         +C_T \sum_{j=1}^{n_T} \xi_{j,k}^t
      \Biggl\} \nonumber \\
    \mathrm{s.t.} \quad
  & y_{i,k}^s (\bm{\theta}_k^T\bm{x}_i^s - b_k) \geq 1 - \xi_{i,k}^s  \nonumber\\
  & y_{j,k}^t (\bm{\theta}_k^T\bm{W} \left({\bm{x}_j^t \atop 1}\right) - b_k) \geq 1 - \xi_{j,k}^t  \nonumber\\
  & \xi_{i,k}^s \geq 0, \quad \xi_{j,k}^t \geq 0
  \label{eq:MMDT-step1} \\
  \min_{\bm{W},\xi^t_{j,k} } \
  & \frac{1}{2}\|\bm{W}\|_F^2 
  + C_T \sum_{j=1}^{n_T} \sum_{k=1}^K \xi_{j,k}^t \nonumber \\
  \mathrm{s.t.} \quad
  & y_{j,k}^t (\bm{\theta}_k^T\bm{W} \left({\bm{x}_j^t \atop 1}\right) - b_k) \geq 1 - \xi_{j,k}^t \nonumber\\
  & \xi_{j,k}^t \geq 0 
  \label{eq:MMDT-step2}
\end{align}

\begin{figure}[t]
  \centering
  \includegraphics[width=.3\linewidth]{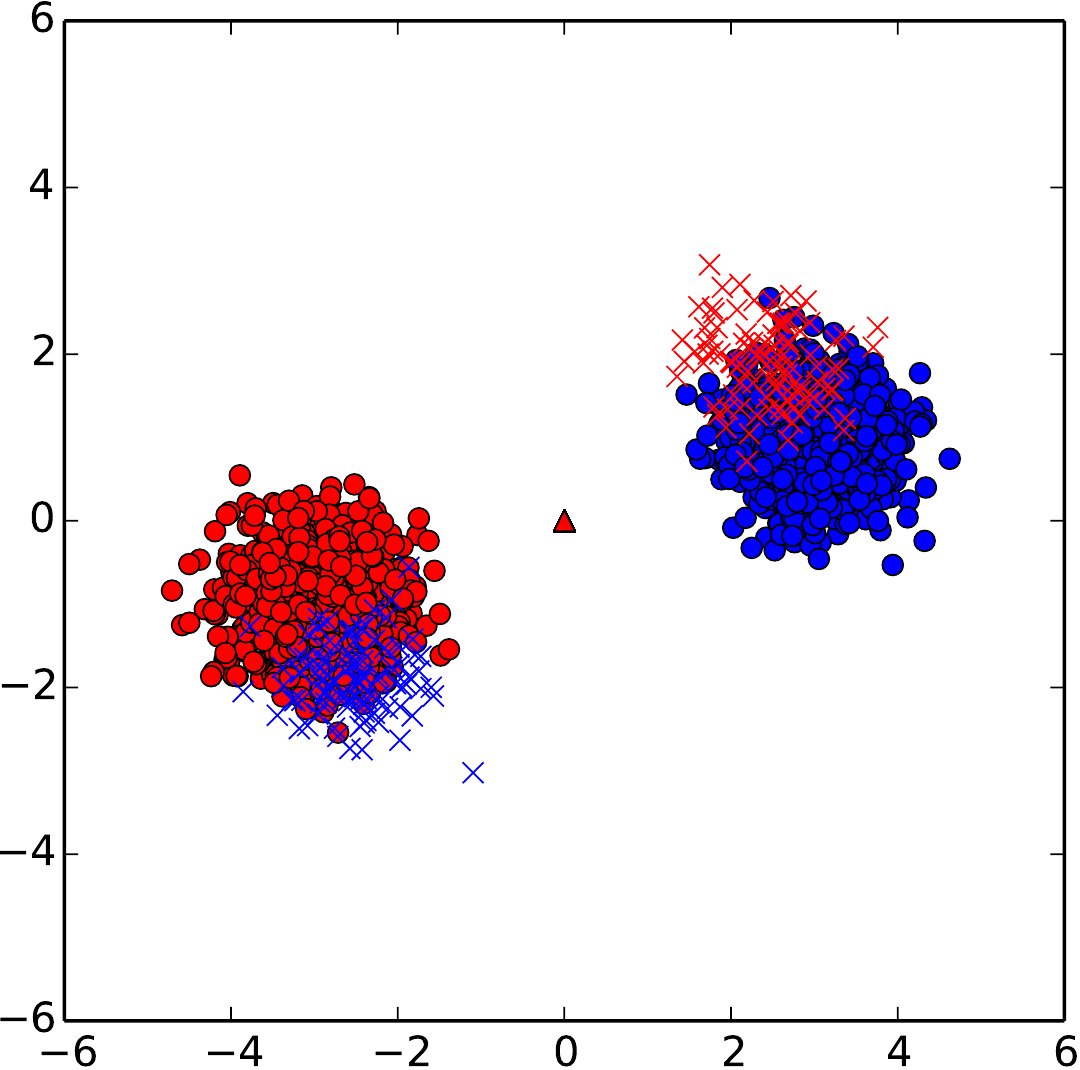}
  \includegraphics[width=.3\linewidth]{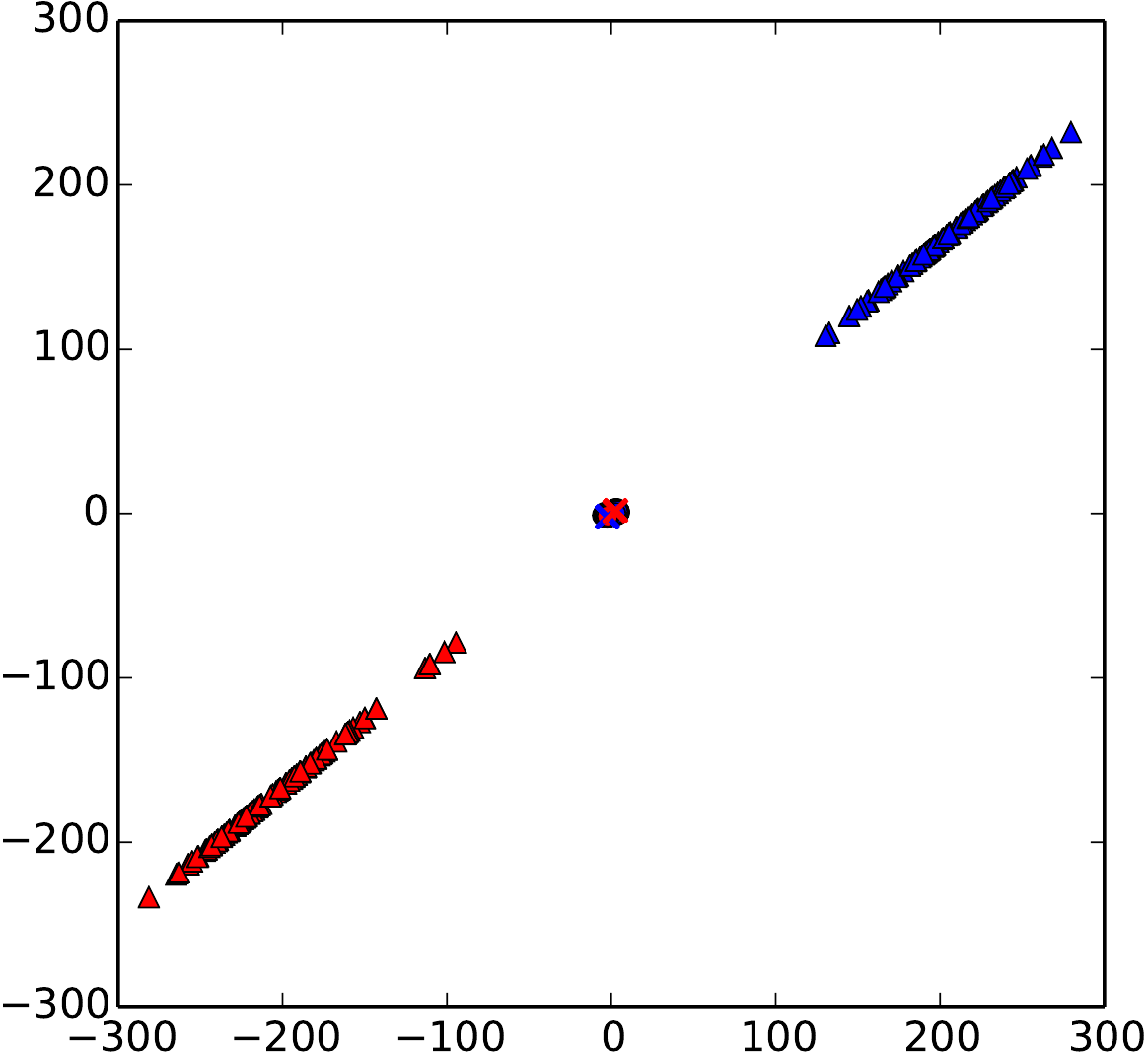}
  \includegraphics[width=.3\linewidth]{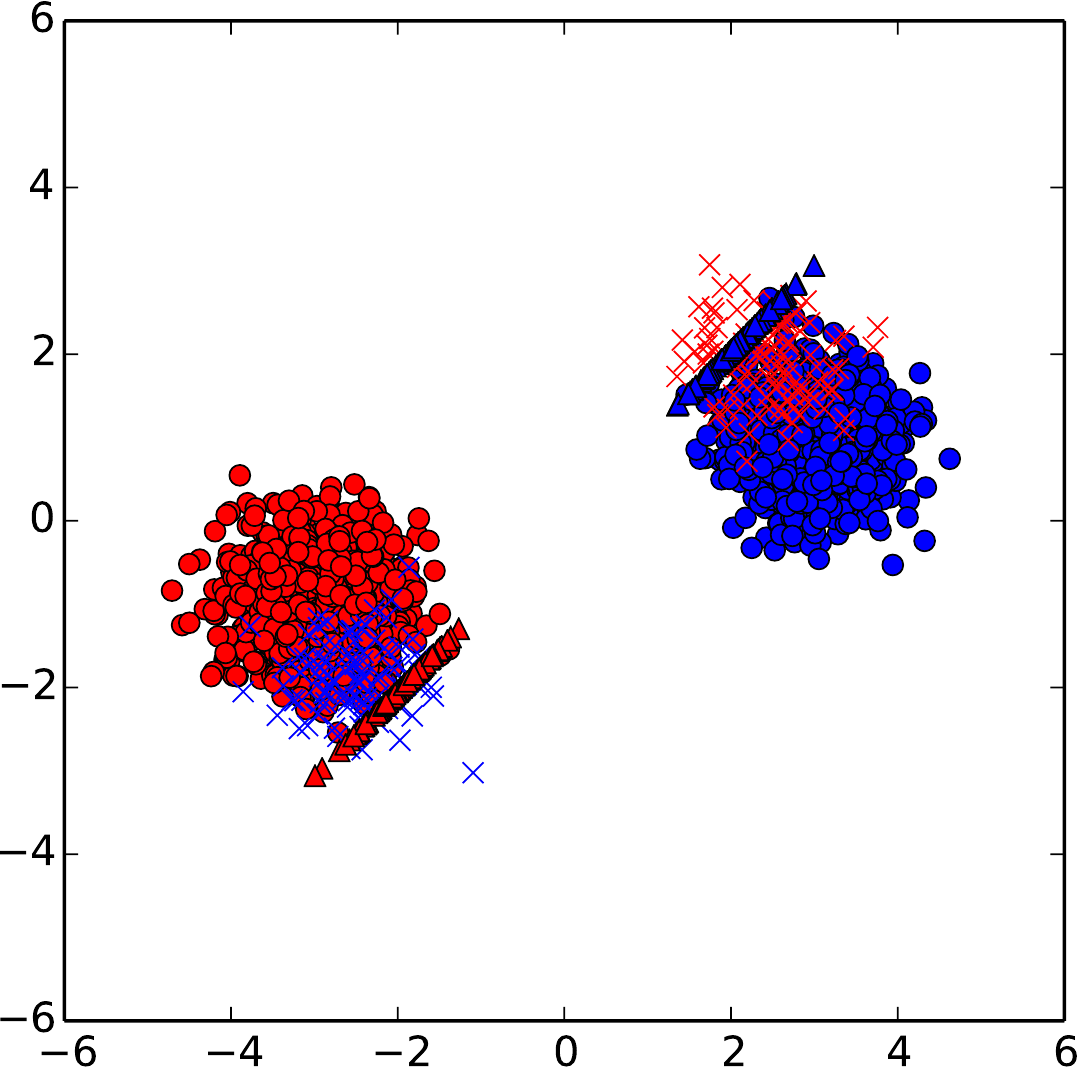}
  \caption{A toy example of a transformation estimated by MMDT and MMDTL2.
  (left) Source and target distributions of two-class problem.
  Red and blue markers are different classes, and
  circles are samples in the source data set,
  and crosses are in the target data set.
  Triangle markers are transformed target samples (here those are all zero because of the initial value of $\bm{W} = \bm{0}$.)
  (middle) The transformed target samples are aligned in a line and distributed far away from source samples.
  (right) The transformed target samples are close to the source distribution.
  }
  \label{fig:mmdt_degenerated}
\end{figure}

A problem of the method is that 
the estimated linear transformation $\bm{W}$ can be degenerated due to minimizing the Frobenious norm.
An example is shown in Figure \ref{fig:mmdt_degenerated} where the target distributions of a spherical shape are
transformed into a collinear distribution, and are not similar to the source distributions.
Because of this effect, the margin after the transformation is smaller than the one between classes of the source distribution.
We show that this is problematic in experimental results.

\section{Proposed method: MMDTL2}

In order to solve the problem of MMDT above, we add the $L_2$ distance constraints
to ensure that the source and transformed target distributions are similar to each other.
The optimization problem for estimating $\bm{W}$ is now formulated as follows:
\begin{align}
  \min_{\bm{W}, \xi^t_{i,k}} \
  & \frac{1}{2}\|\bm{W}\|_F^2 + C_T\sum_{i=1}^{n_T}\sum_{k=1}^K\xi_{i,k}^t \nonumber \\
  & +\frac{1}{2}D\sum_{i=1}^{n_T}\sum_{j=1}^{n_S}y_{i,j}\|(\bm{W} \left({\bm{x}_j^t \atop 1}\right) -\bm{x}_j^s)\|_2^2 
  \label{eq:Proposed_obj} \\
  \mathrm{s.t.} \quad
  & y_{i,k}^t (\bm{\theta}_k^T \bm{W} \left({\bm{x}_j^t \atop 1}\right) - b_k ) \geq 1 - \xi_{i,k}^t \nonumber \\
  & \xi_{i,k}^t \geq 0, \nonumber 
\end{align}
where $D$ is a scaler, and $y_{i,j}$ is a weight between $\bm{x}_i^t$ and $\bm{x}_j^s$.

For simplicity, hereafter we restrict that the last column of $\bm{W}$ is zero, in other words, $\bm{W} \in R^{M_S \times M_T}$ and all forms of $\bm{W}\left({\bm{x}_j^t \atop 1}\right)$ become $\bm{W} \bm{x}_j^t$.
Then by rewriting the problem we have the following standard form of quadratic programing:
\begin{align}
  \min_{\bm{w}, \xi^t_{i,k}} \ 
  & \frac{1}{2} \| \bm{w} \|_2^2 + C_T\sum_{k=1}^K\sum_{i=1}^{n_T}\xi_{i,k}^t 
  \label{eq:Proposed_obj2} \\
  & +\frac{1}{2} D\sum_{i=1}^{n_T} \sum_{j=1}^{n_S}y_{i,j} \Bigr( \bm{w}^TU(\bm{x}_i^t) \bm{w}  - 2 \bm{w}^T \bm{v}_{i,j} + \| \bm{x}_j^s \|_2^2 \Bigl) 
         \nonumber \\ 
  \mathrm{s.t.} \quad
   & \xi_{i,k}^t \geq 0, \nonumber \\
   & y_{i,k}^t (\bm{\phi}_k^T(\bm{x}_i^t)\bm{w} - b_k) \geq 1 - \xi_{i,k}^t \nonumber 
\end{align}
where $\bm{w}=\mathrm{vec}(\bm{W}) \in R^{M_S M_T}$ and $\bm{\phi}_k(\bm{x})=\mathrm{vec}(\bm{\theta}_k \bm{x}^T) \in R^{M_S M_T}$
so that $\bm{\theta}_k^T \bm{W} \bm{x}_i^t = \bm{\phi}_k^T(\bm{x}_i^t)\bm{w}$,
and
\begin{align}
  U(\bm{x}) &= \left[
  \begin{array}{cccc}
  \bm{x} \bm{x}^T & & & \\
  & \bm{x} \bm{x}^T & & \\
  & & \ddots & \\
  & & & \bm{x} \bm{x}^T 
  \end{array}
  \right] \in R^{M_S M_T \times M_S M_T}
  \label{eq:def_U} \\
  \bm{v}_{i,j} &= \mathrm{vec}(\bm{x}_j^s (\bm{x}_i^t)^T) \in R^{M_S M_T}.
  \label{eq:def_v}
\end{align}
This quadratic programming however is not efficient to solve
because the number of variables is $M_S M_T + n_T$,
therefore solving this quadratic programming 
requires about $\mathcal{O}( (M_S M_T + n_T)^3)$ computations (e.g., \cite{Yinyu1989}).
This means that the primal problem becomes intractable quickly as the feature dimension increases.
Next we therefore derive the dual problem to this primal problem in order to reduce the computation cost.

To this end, we first obtain the Lagrangian 
\begin{align}
   L 
   &= \frac{1}{2} \|\bm{w}\|_2^2 +C_T\sum_{k=1}^K\sum_{i=1}^{n_T}\xi_{i,k}^t 
   - \sum_{k=1}^K\sum_{i=1}^{n_T}\mu_i\xi_{i,k}^t 
   \label{eq:Proposed_AG} \\
   & - \sum_{k=1}^K \sum_{i=1}^{n_T} a_{i} 
      (y_{i,k}^t (\bm{\phi}_k^T(\bm{x}_i)\bm{w} - b_k) - 1 + \xi_{i,k}^t )
    \nonumber \\
   & +\frac{1}{2}D\sum_{i=1}^{n_T}\sum_{j=1}^{n_S}y_{i,j}(\bm{w}^T\bm{U}(\bm{x}_i^t)\bm{w}
    - 2 \bm{w}^T \bm{v}_{i,j}
   + \| \bm{x}_i^s \|_2^2 ), \nonumber
\end{align}
where $a_i$ and $\mu_i$ are Lagrangian multipliers.
Then we take the derivatives with respect to $\bm{w}$ and $\xi_{i,k}^t$ and let them to zero;
\begin{align}
  \frac{\partial L}{\partial \bm{w}} 
  &= \bm{w} -\sum_{k=1}^K\sum_{i=1}^{n_T}a_iy_{i,k}^t\bm{\phi}_k(\bm{x}_i^t) \nonumber \\
  &+D\sum_{i=1}^{n_T}\sum_{j=1}^{n_S}y_{i,j}\left( \bm{w}^TU(\bm{x}_i^t)-\bm{v}_{i,j} \right) = 0 
  \label{eq:diff_w} \\
  \frac{\partial L}{\partial \xi_{i,k}^t} 
  &= C_T-a_i-\mu_i = 0
  \label{eq:diff_xi}
\end{align}
After that, we substitute Eqs. (\ref{eq:diff_w}) and (\ref{eq:diff_xi}) into Eq. (\ref{eq:Proposed_obj2}),
and rearrange the equation with respect to $a_i$ then finally we have 
the following quadratic programming.
\begin{align}
  \max_{a_1,a_2,\ldots}
  & \frac{-1}{2}
    \sum_{i=1}^{n_T} \sum_{j=1}^{n_T} 
      a_i a_j 
      \sum_{k_1=1}^K \sum_{k_2=1}^K 
      y_{i,k_1}^t y_{j,k_2}^t 
      \bm{\phi}_{k_1}^T(\bm{x}_i^t)
      \bm{V}^{-1}
      \bm{\phi}_{k_2}(\bm{x}_j^t)
    \nonumber \\
  & + \sum_{k=1}^K \sum_{i=1}^{n_T}
    a_i 
    \left( 
    1  
    - D \bm{\phi}_k^T(\bm{x}_i^t) \bm{V}^{-1}
          \sum_{m=1}^{n_T} \sum_{n=1}^{n_S} y_{m,n}
  \bm{v}_{m,n} 
  \right) 
  \label{eq:Proposed_obj_dual}
\end{align}
Later we compute $\bm{w}$ by using estimated $a_i$ and Eq. (\ref{eq:atow}) as follows:
\begin{align}
  \bm{w} &=
   \bm{V}^{-T} 
   \left(
    \sum_{k=1}^K \sum_{i=1}^{n_T} 
      a_i y_{i,k}^t \bm{\phi}_k(\bm{x}_i^t)
      + D \sum_{i=1}^{n_T} \sum_{j=1}^{n_S} y_{i,j} \bm{v}_{i,j} 
   \right) 
  \label{eq:atow}
\end{align}
where
\begin{align}
  \bm{V} = \bm{I}+D\sum_{i=1}^{n_T}\sum_{j=1}^{n_S}y_{i,j}\bm{U}(\bm{x}_i^t).
  \label{eq:V}
\end{align}
This dual formulation is much smaller than the primal problem
because the number of variables $a_i$ is now $n_T$ which is much smaller than $M_S M_T$ in general.

\section{Experimental results}

In this section, we show two kind of experimental results.
First, we compare the computation cost between the primal and dual problem formulations.
Second, we compare the proposed MMDTL2 with MMDT.

We use two different datasets. The source data set has 400 NBI images
(Type A: 200, Type B,C3: 200) taken by OLYMPUS EVIS LUCERA endoscopy \cite{EVIS_LUCERA_H260AZL},
and the target data set has 180 NBI images (Type A: 90, Type B,C3: 90)
taken by newer OLYMPUS EVIS LUCERA ELITE endoscopy \cite{EVIS_LUCERA_ELITE_HQ290Z}.

\subsection{Comparison between primal and dual}

Table \ref{tab:ComputationTime_depth6} show computation time of the primal and dual formulations, where the feature dimension is $M_S = M_T = 128$.
The setup time counts how long it takes to compute coefficients of $\bm{w}$ in Eq. (\ref{eq:Proposed_obj2}) for the primal,
or coefficients of $a_i$ in Eq. (\ref{eq:Proposed_obj_dual}) for the dual.
The optimization time is the computation time to solve a quadratic program,
the calculation time is to compute $\bm{w}$ from $a_i$, meaningful only for dual.
In this setting, the dual formulation of MMDTL2 is about 12 times faster than the primal.


\begin{table}[t]
  \centering
  \caption{Computation time (in second) of MMDTL2.}
  \label{tab:ComputationTime_depth6}
  \begin{tabular}{c| cc}
& primal & dual \\ \hline
setup & 402.90 & 538.18 \\
optimization & 6396.84 & 0.060 \\
calculation & N/A & 0.023 \\ \hline
total & 6799.74 & 538.26 
  \end{tabular}
\end{table}

\subsection{Performance comparison}

\begin{figure}[t]
  \centering
  \includegraphics[width=.6\linewidth]{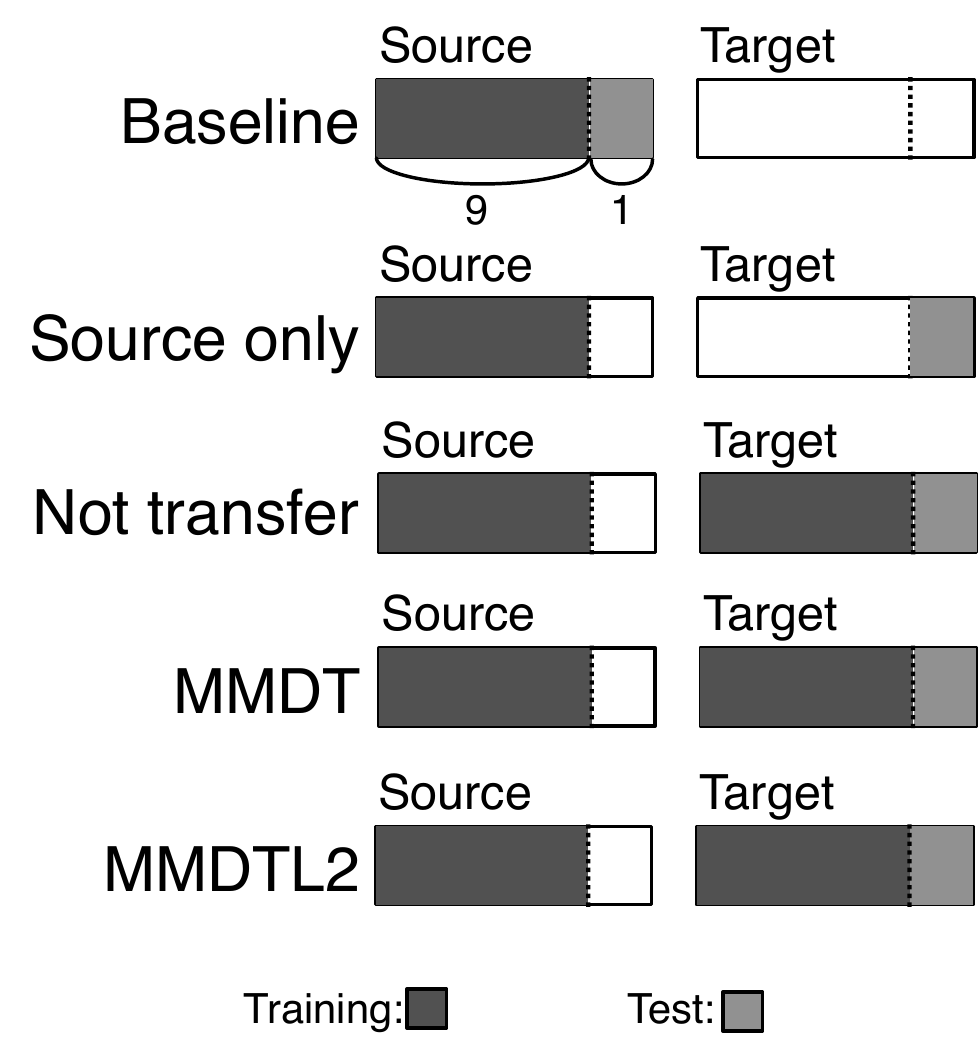}
  \caption{Performance with different experiment settings.}
  \label{fig:Exp_env}
\end{figure}

In this experiment, we compare performances of the proposed method with the following settings, which are similar to \cite{sonoyama2015transfer}.
These settings are shown in Figure \ref{fig:Exp_env}
in terms of 10-fold cross validation.

\begin{description}

\item[Baseline]\mbox{}\\
For comparison, we first perform an experiment without any transfer learning by using the source dataset only.
In this case, the source dataset is divided into 10 folds;
a normal 10-fold cross validation with the source dataset.

\item[Source only]\mbox{}\\
The second setting also doesn't use any transfer learning but with the source and target datasets.
This means that source images are used for training, and the resulting classifier is simply used to classify target (test) images. We divide the source and target datasets in to 10 folds (each source fold in the source corresponds to a target fold). Then 9 source folds are used for training, and the remaining one target fold is used for testing.

\item [Not transfer]\mbox{}\\
The third setting doesn't use any transfer learning, too. Target images are however also used for training unlike the previous settings above.
We divide the source and target datasets in to 10 folds,
then 18 folds (9 source and 9 target folds) are used for training a classifier. The remaining target fold is used for testing by simply applying the trained classifier.
Note that the number of training images is now doubled while the number of testing images remains the same.

\item[MMDT]\mbox{}\\
Fourth setting is MMDT, the existing method.
We use 9 source and 9 target folds for estimating the linear transformation $\bm{W}$ and classifier $\bm{\theta}$.
For training, those 18 folds are used. Features $\bm{x}^t$ in the 9 target folds are transformed by $\bm{W}$ as $\bm{W}\bm{x}^t$.
The remaining target fold is used for testing, however features in it are transformed by $\bm{W}$ as in the training, and then the trained classifier is applied.

\item[MMDTL2]\mbox{}\\
The last setting is MMDTL2, the proposed method. The setting is the same with MMDT.

\end{description}

\begin{figure}[t]
  \centering
  \includegraphics[width=\linewidth]{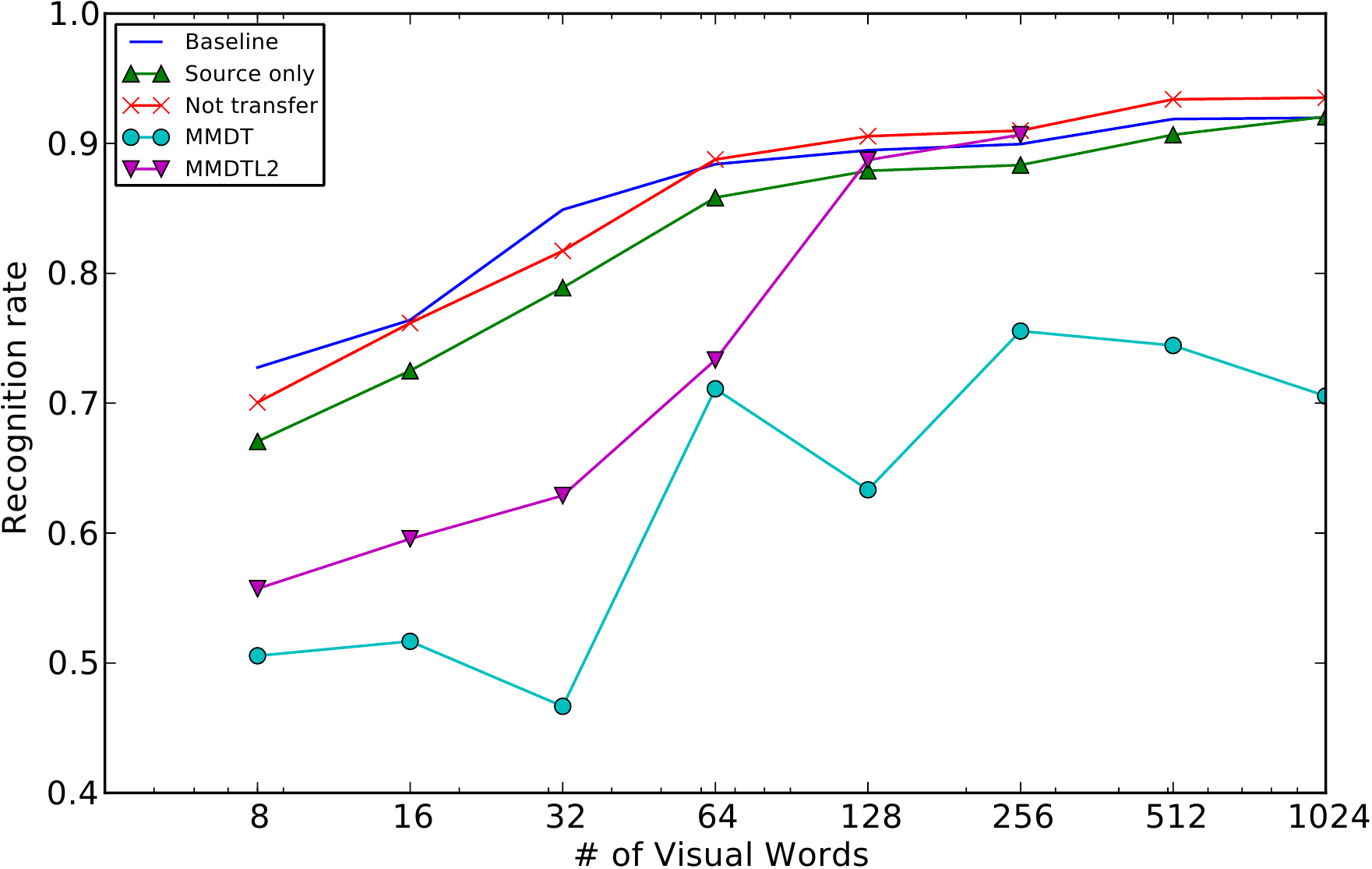}
  \caption{Performance with different experiment settings.
  Horizontal axis is the number of features $M_S = M_T$. }
  \label{fig:mix_result_tr}
\end{figure}

Figure \ref{fig:mix_result_tr} show performances of the different setting.
As expected, the source only setting is slightly worse than the baseline as the distributions of the source and target data sets are different actually.
Results of MMDT are worse in any cases. It might be due to the problem that estimated linear transformations are degenerated.
The proposed MMDTL2 is much better than MMDT because of the additional $L_2$ constraints, and comparable to the baseline in higher dimension.
Also MMDTL2 is expected to behave better than the not transfer setting 
when $M_S = M_T \ge 256$.

\section{Conclusions}

We have proposed the MMDTL2, which extends MMDT by adding $L_2$ distance constraints, for transfer learning of medical images.
Also we have derived the dual formulation of the quadratic programming in order to achieve smaller computation cost. Still the proposed method needs a high computation cost for preparing the quadratic programming, therefore we will reduce further the computation cost so that feature vectors in much larger dimension can be handled.

\bibliographystyle{ieeetr}
\bibliography{bib,ref_hirakawa/endoscopy,ref_hirakawa/related_work,ref_hirakawa/image_classification,ref_hirakawa/hafner}

\end{document}